\documentclass[11pt,a4paper]{article}
\usepackage[hyperref]{emnlp-ijcnlp-2019}
\usepackage{times}
\usepackage{latexsym}

\usepackage{url}
\usepackage{makecell}
\usepackage{multirow}
\usepackage{amsmath, amssymb}
\usepackage{tabularx}

\usepackage[final]{graphicx}
\graphicspath{{figs/}}

\aclfinalcopy % Uncomment this line for the final submission
\DeclareMathOperator*{\argmax}{\arg\!\max}
\DeclareMathOperator*{\E}{\mathbb{E}}

\newcommand{\qv}{{\bf q}}
\newcommand{\tv}{{\bf t}}
\newcommand{\pv}{{\bf p}}

\newcommand\tsup[1]{\textsuperscript{#1}}

\newcolumntype{L}[1]{@{}>{\centering\hsize=#1\hsize\raggedright\arraybackslash}X@{}}
\newcolumntype{R}[1]{@{}>{\centering\hsize=#1\hsize\raggedleft\arraybackslash}X@{\,}}% X@{}
\newcolumntype{C}[1]{@{}>{\centering\hsize=#1\hsize\centering\arraybackslash}X@{}}%

\title{Subword Language Model for Query Auto-Completion}

\author{
    Gyuwan Kim  \\
    Clova AI, NAVER Corp. \\
    {\tt gyuwan.kim@navercorp.com}
}

\date{}

\begin{document}

\maketitle

\begin{abstract}
Current neural query auto-completion (QAC) systems rely on character-level language models, but they slow down when queries are long. We present how to utilize subword language models for the fast and accurate generation of query completion candidates. Representing queries with subwords shorten a decoding length significantly. To deal with issues coming from introducing subword language model, we develop a retrace algorithm and a reranking method by approximate marginalization. As a result, our model achieves up to 2.5 times faster while maintaining a similar quality of generated results compared to the character-level baseline. Also, we propose a new evaluation metric, mean recoverable length (MRL), measuring how many upcoming characters the model could complete correctly. It provides more explicit meaning and eliminates the need for prefix length sampling for existing rank-based metrics. Moreover, we performed a comprehensive analysis with ablation study to figure out the importance of each component\footnote{Code is available at \url{https://github.com/clovaai/subword-qac}.}.
\end{abstract}

\section{Introduction}

%%% What is QAC? Why is it necessary and helpful?
Query auto-completion (QAC) is one of the essential features for search engines.
When a user types a query in the search box, QAC systems suggest most likely completion candidates \cite{cai2016survey}.
It not only saves time for users to enter search terms but also provides new information more than what was initially expected.
%by showing that the search results are different from

%%% Limitation of character-level model: long sequence -> slow and inaccurate (error propagation, large search space)
Recent neural QAC models in the literature employ character-level language models \cite{park2017neural}.
It is a natural choice in that QAC systems need to respond whenever a user enters a query as input character-by-character.
In addition to the accuracy, speed in terms of latency is also an indispensable prerequisite for practical QAC systems.
%%% Latency 
% Inference of the model should be fast in latency, but longer sequence length slows down the inference.
The generation process is auto-regressive, and the size of the search space is exponential to the sequence length. 
Long character sequences make prediction slow and inaccurate in the constraints of limited computation.
Also, character-level models are prone to errors due to long-range dependency \cite{sennrich2016grammatical}.
Therefore, these limitations arouse to consider alternatives to represent a query in a shorter sequence.

%%% Subword language model
In this paper, we apply a subword language model for query auto-completion.
Compared to character language models, subword language models reduce sequence length and the number of decoding steps significantly, thus resulting in much faster decoding.
%%% Subword segmentation
For subword-level modeling, a segmentation algorithm is necessary.
Byte pair encoding (BPE) \cite{sennrich2015neural} is widely used, but noise in the data makes segmentation ambiguous and degrades BPE output.
To address this issue, as well as BPE, we use subword regularization (SR) algorithm proposed by \citet{kudo2018subword} that stochastically samples multiple segmentations by utilizing a unigram language model.
To our knowledge, we are the first to apply SR to language modeling.
%, instead of encoder-decoder models.
% We use these segmentation algorithms (BPE, SR) for training a subword language model and decoding with the trained model.

Interestingly, language models for QAC should take care of the last token that may be incomplete.
Like character language models, subword language models can represent incomplete tokens because it can generate any subsequence of sentences, whereas word language models cannot.
%%% Additional techniques
If we segment prefix as given to encode it using neural networks, the segmentation of prefix may not match with that of ground truth query because the prefix is an incomplete substring of the original desired query.
In that case, this enforced segmentation is less likely to appear in training, especially for deterministic segmentation such as BPE.
As a result, the model starting from this segmentation is unlikely to generate ground truth query.
To consider every possible segmentation of target completion, we propose retrace algorithm that is going a few characters back from the end and generating candidates with the restriction that they should match with retraced characters.
%%% Reranking with approximate marginalization
For the case of SR models, due to the stochasticity of segmentation, we should marginalize over all possible segmentations to calculate the likelihood of a query.
For better approximation than just $\argmax$, we perform reranking with approximated marginalization using the output of beam search.
Experimental results show that these techniques improve the robustness of the decoding process of the subword language model to achieve close generation quality compared to the character baseline.
% Also, we examine $n$-best decoding, which means starting decoding with $n$ segmentations of the prefix.

%%% Evaluation metric
We propose a novel metric for query auto-completion evaluation, called mean recoverable length (MRL).
This metric remedies shortcomings of common QAC evaluation metrics, mean reciprocal rank (MRR) and partial-matching MRR (PMRR), which require sampling of a prefix length and are favorable to short queries.
We conduct comprehensive ablation study and analysis of our models on these three metrics.

\section{Related Work}
\label{related_work}

%%% QAC system overview: traditional vs. neural 
One of the successful traditional QAC approaches is most popular completion (MPC) \cite{bar2011context}, which returns the most frequent candidates among all previously observed queries that match the prefix.
After extracting candidates, reranking algorithms (e.g., LambdaMART \cite{burges2010ranknet}) with additional features are used to align final candidates.
These methods cannot generate previously unseen queries by nature.
Contrary to traditional approaches based on information retrieval, neural approaches can generalize to unseen prefixes.

%%% Granularity of NLP models %, character-based
Choosing an appropriate granularity level for text segmentation has been long studied over the variety of natural language processing problems.
It can be a character, subword, word, phrase, sentence, and even paragraph.
A trade-off between them exists, and the best performing granularity often varies depending on tasks and datasets.
Character models are widely used to address natural language processing tasks including text classification \cite{kim2014convolutional, zhang2015character, conneau2016very}, language modeling \cite{hwang2017character, al2018character}, machine translation \cite{chung2016character, lee2017fully}, etc.

%%% Subword NMT, subword language modeling
Currently, neural machine translation systems widely use subword segmentation as \textit{de facto}.
\citet{mikolov2012subword} observed that a subword language model is advantageous in that it achieves better performance compared to character-level models with zero out-of-vocabulary rate and smaller model size.
BERT \cite{devlin2018bert} uses a subword as the unit token for their (masked) language models.

%%% Word-level
% Words are regarded as basic units having their explicit meaning in natural language processing.
Word-level segmentation can easily shorten sequence length compared to character-level.
However, word-level models require larger vocabulary size and the number of parameters to learn.
Also, it causes data sparsity issue.
Because the vocabulary of words is usually fixed before training, it cannot generate out-of-vocabulary words by itself.
Search systems are especially in an open vocabulary setting.
%%% Incomplete token
For word-level models, it is hard to deal with the last incomplete token because it may not be in the vocabulary, unlike character-level naturally handle it.
Even if the vocabulary contains this token, the decoding process may be somewhat different from expected.

%%% Word-character hybrid model
Word-character hybrid models were proposed to overcome the out-of-vocabulary problem \cite{luong2016achieving, wu2016google}.
A word-level decoder generates a word sequence, and when it generates a special \verb <UNK>  token, a character-level decoder generates a character sequence on top of it.
These two decoders are connected hierarchically.
Word models assume whitespace as the boundary of words. 
In some languages including Japanese and Chinese, segmentation of sentences into words is unknown in advance and sometimes vague. % not readily given
Moreover, input queries usually include much noise such as typos, grammatical errors and spacing errors.
The problems mentioned above hinder word-level processing for QAC. % This noisiness

%%% Word information
\citet{park2017neural} and \citet{fiorini2018personalized} incorporate word information by concatenating its embedding with character embedding only at the word boundary and use a special \verb <INC>  token embedding for non-boundary positions.
This mechanism is inefficient in that the word signal is sparse. % given sparsely. % naive and
Most of the word-character hybrid models focus on input representation rather than generation.
Usually, their representations are concatenated, or composition functions are learned \cite{kim2016character, miyamoto2016gated}.
Even though they use word information to the input, the decoding process of their models is still in the character-level.

%%% Non-autoregressive decoding
We can interpret generating a subword which is a concatenation of characters as parallel decoding of characters \cite{stern2018blockwise}.
In this sense, non-autoregressive neural machine translation \cite{gu2017non, lee2018deterministic} is related to our work.
They also aim to improve decoding speed with minimal performance degradation.
%We believe that o
Our model and decoding method can be used for non-autoregressive NMT in place of a character-level decoder, and in the opposite direction, we can apply their approaches to QAC vice versa.

\section{Subword Language Model}

%We describe our model formally.
Let $\Sigma$ be the set of all possible characters and $V$ be the vocabulary of tokens. 
% $\subseteq \Sigma^{*}$ \footnote{For arbitrary  $S^{*}$ means a set of variable length sequences whose elements are in $S$. For example, $\{a, b\}^{*} = \{ a, b, aa, ab, ba, bb, aaa, \cdots\}$. $S^{*}_{+}$ equals to $S^{*}$ and $S^{*}_{0}$ includes an empty string in addition to $S^{*}$.}
Each token is a character or a concatenation of characters, and it is a subword in our case.
A language model estimates the probability of a token sequence $\tv$ where the probability distribution of token $t_i$ at each step $i$ is conditioned on the previous tokens $t_{<i}$:
\begin{equation*}
  p(\tv;\theta) = \prod_{i=1}^{|\tv|} p(t_i|t_{<i};\theta)
\end{equation*}
where $t_i \in V$ and $\theta$ is a set of model parameters.
For a token sequence $\tv$, we can map it to a query $\qv = concat(\tv) (= t_1 \oplus t_2 \oplus \cdots \oplus t_{|\tv|})$ by concatenating itself sequentially.
Then, the probability of a given query $\qv$ is the sum of the probability over the set of all possible segmentation $\tv$, $S(\qv) = \{\tv: concat(\tv) = \qv \}$: %\subseteq
\begin{equation*}
  p(\qv;\theta) = \sum_{\tv\in S(\qv)} p(\tv;\theta).
\end{equation*}
% $x = (x_1, \cdots, x_{|x|})$ ($x_i \in \Sigma$)
%where $S(\qv)$ is the set of possible segmentations of $\qv$.
% $S(\qv)$ depends on the segmentation algorithm. %: in our case, BPE or SR. -> no
Similar to \cite{chan2016latent}, segmentation $\tv$ can be interpretable as a latent decomposition of the query $\qv$.
%As pointed out by \cite{chan2016latent}, segmentation can be interpreted as a latent decomposition of the query.

\subsection{Segmentation}
% We use a character-level language model as a baseline for our proposed model.
In character-level language modeling, token vocabulary $V$ is equal to $\Sigma$, and segmentation is performed by merely splitting every character.
% In the word-level language modeling, we split a sentence by whitespace.
We exclude word-level language modeling which splits a sentence by whitespace from consideration due to its limitations mentioned in Section \ref{related_work}.

In the case of subword language modeling, we use two widely used segmentation algorithms: (1) byte pair encoding (BPE) and (2) subword regularization (SR).
Formally, a segmentation algorithm defines a probability distribution over a token sequence $\tv$ conditioned on given query $\qv$: $p_{seg}(\tv | \qv)$.

The BPE algorithm is deterministic because it segments greedily from left to right. % to achieve its unique segmentation. %so that one-to-one correspondence between a sentence and its segmentation holds.
% i.e. $|S(\qv)|=1$ for any query $\qv$.
On the other hand, SR can sample multiple segmentations stochastically.
The number of possible segmentations is exponentially large.
It is hard to calculate the likelihood of a given sentence using dynamic programming because even with the same prefix, hidden states vary upon different previous tokenization.
Marginalization over all possible segmentations of very long sequences is intractable.
In sum, we compare character-level and subword-level (BPE, SR) language modeling.

\subsection{Training} \label{sec:training}
We can derive an unbiased gradient estimator of the log-likelihood of a query by using Bayes' theorem and the identity of $\nabla_{\theta} f(x; \theta) = f(x; \theta) \nabla_{\theta} \log f(x; \theta)$ assuming $f(x; \theta) \neq 0$ for all $x$ \cite{williams1992simple}:
\begin{equation*}
    \nabla_{\theta} \log p(\qv; \theta) = \E_{\tv \sim p(\tv | \qv ; \theta)} \nabla_{\theta} \log p(\tv; \theta).
\end{equation*}
% \partial
However, since sampling $\tv$ from $p(\tv | \qv ; \theta)$ is computationally expensive, we heuristically use $p_{seg}(\tv | \qv)$ instead. % in a heuristic way. 
Regardless of the language model parameters $\theta$, segmentation model $p_{seg}$ is learned before the language model training and can be used to sample $\tv$ easily.
The better way to approximate the distribution $p(\tv | \qv ; \theta)$ will be explored in the future.

Our training objective becomes equivalent to maximizing the average log-likelihood of the segmentation of sentences:
\begin{equation*}
  \begin{aligned}
    L(\theta) &= \frac{1}{|Q|} \sum_{\qv \in Q} \log p(\tv;\theta) \\
              &= \frac{1}{|Q|} \sum_{\qv \in Q} \sum_{i} \log p(t_i|t_{<i};\theta),  
    % \E_{\tv \sim P(\tv|\qv)}
  \end{aligned}
\end{equation*}
where $Q$ is the training set of all queries, and $\tv$ is the segmentation of a query $\qv$ sampled from $p_{seg}(\tv | \qv)$ which depends on the segmentation algorithm.
This objective is equal to the average negative log-likelihood of sentences if and only if the segmentation is deterministic.
% For a character-level model and BPE segmentation, segmentation is deterministic, $|S(\qv)| = 1$.
% For SR, $\tv$ is sampled from $\qv$.
The gradients of the loss function are computed using the back-propagation through time (BPTT) \cite{rumelhart1986learning}.

\section{Decoding} \label{sec:decoding}

Given prefix $\pv$, let the set of its completions be $Q_{+}(\pv)$ and the set of their tokenizations be $S_{+}(\pv) = \{\tv: concat(\tv) \in Q_{+}(\pv)\}$.
% =\pv \times \Sigma^{*}_{+}
We want to find the most likely completion $\hat{\qv}$:
\begin{align}
    \hat{\qv} &= \argmax_{\qv} \log p(\qv|\pv) \nonumber \\
              &= \argmax_{\qv \in Q_{+}(\pv)} \log p(\qv) \nonumber \\
              &= \argmax_{\qv \in Q_{+}(\pv)} \log \sum_{\tv \in S(\qv)} p(\tv) \label{exact_decoding}
\end{align}
% label smoothing -> sharp -> approximation is more correct
however, this is obviously intractable to search in the infinitely large token sequence space.
We approximate this by decoding for the best token sequence $\hat{\tv}$ and then returning its corresponding query $\tilde{\qv}$ by concatenating its token sequentially:
\begin{align*}
  \hat{\tv}   &= \argmax_{\tv \in S_{+}(\pv)} \log p(\tv)\\
  \tilde{\qv} &= concat(\hat{\tv})
\end{align*}
% For the case of deterministic segmentation where $S(\qv) = 1$, $\tilde{\qv}$ is exactly same with $\hat{\qv}$, but this is not guaranteed when multiple segmentations exist.  % <- not true (q->t deterministic but t-> q can be a lot)

Basic practice is segmenting $\pv$, feeding it into language model to encode $\pv$, and using it for the decoding.
Since finding $\hat{\tv}$ is also intractable, beam search decoding is used but only results in suboptimal predictions.
We will improve this incrementally with techniques following.

\subsection{Retrace Algorithm}
\begin{figure}[t!]
    \centering
    \includegraphics[width=0.45\textwidth]{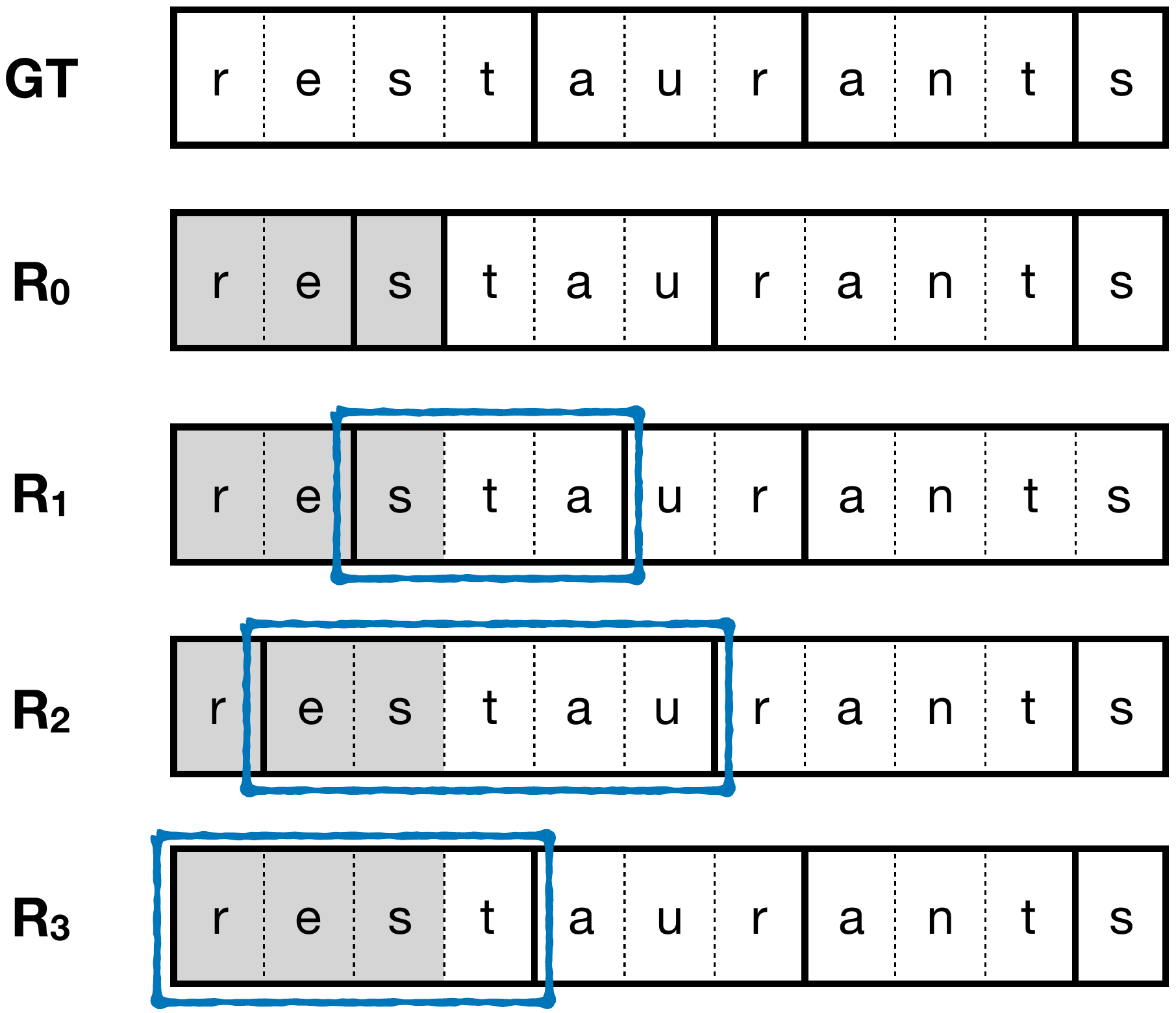}
    \caption{
        \label{fig:retrace}
        Illustration of retrace algorithm with the example of ``restaurants.'' The gray area means given prefix (``res") of the query. The solid line indicates the boundary of the segmentation. GT is the segmentation of ground truth query. Possible examples of the generated sequence of tokens belonging to the case $R_r$ are visualized. Blue boxes indicate a fixed segmentation with retrace algorithm at the end of the prefix. 
    }
\end{figure}

There is no guarantee that the end of given prefix matches the tokenization boundary of the completed query.
To address this possibility of the incompleteness at the end of a prefix, we can retrace a few characters and generate from there.
For the case (call it $R_r$) where the last token that overlaps with the prefix finishes $r$ characters before the end of the prefix, first starting from a token sequence of $\pv_{1:|\pv|-r}$, we can perform beam search decoding on the restriction that the next token should cover the remaining part of the prefix and the next new character.
Figure \ref{fig:retrace} illustrates this algorithm.

%For a token sequence $\tv$ in $S_{+}(\pv)$, let the position of the last token that overlap with the prefix is $r$ characters before the end of the prefix. More precisely, there exist $e$ where $concat(t_{1:e}) = \pv_{1:|\pv|-r}$, $|t_{e+1}| > r$, and $t_{e+1}$
% Next token should overlap and continue from the remaining part of the prefix.
%$S_{+}(\pv)$ is partitioned into sets of segmentation:
%\begin{align*}
%    R_r &= S(\pv_{1:|\pv|-r}) \times V_r \times V^{*}_{0} \\
%    V_r &= \{t: t \in V, |t| > r, t_{1:r} = \pv_{|\pv|-r+1:|\pv|} \},
%      % t: single token
%\end{align*}
%for $0 \leq r \leq |\pv|$.

This process is unnecessary for a character-level model since every token is a single character.
On the other hand, the retrace algorithm is helpful for subword models, especially BPE models which have deterministic segmentation algorithm.
% no overlap b/c different hidden

We can limit the maximum step of retrace by $L$ to only consider $R_{r}$ where $0 \leq r \leq L$ because of the computational issue.
We will denote this limitation as $R^{L}$.
$R^0$ is the usual case without the retrace algorithm, and $R^{\infty}$ counts every possible retrace steps.

\subsection{Reranking Method by Approximate Marginalization}
QAC system has to suggest $N$ completion candidates sorted in order of likelihood rather than finding only the best completion candidate.
We can extract a set of top $B (\geq N)$ candidates using beam search with beam size $B$, namely $T_B = \{\tv^1, \cdots, \tv^B\}$ in the descending order of likelihood.
In the case of deterministic segmentation, $\qv^i = concat(\tv^i)$ are mutually different. i.e. $|Q_B|=B$ for $Q_B=\{\qv^1, \cdots, \qv^B\}$.
Then, trivially our prediction would be $(\qv^1, \cdots, \qv^N)$.

On the other hand, in the case of stochastic segmentation, same query $\qv^i, \qv^j (i \neq j)$ with different token sequence $\tv^i, \tv^j$ may exist.
The obvious way is merely removing duplicates.

On the assumption that $\log p(\tv) \gg \log p(\tv')$ for all $\tv \in T_B$ and $\tv' \notin T_B$, Equation \eqref{exact_decoding} implies that marginalization over the final beam outputs can provide better approximation:
\begin{equation*}
  \begin{aligned}
    \hat{\qv} &\approx \argmax_{\qv \in Q_{+}(\pv)} \log \sum_{\tv \in T_B} p(\tv) \\
              &= \argmax_{\qv \in Q_B} \log \sum_{\tv \in T_B} p(\tv)
  \end{aligned}
\end{equation*}
In other words, reranking after summing out the probability of duplicates can give better ordered list of candidates.

%\subsection{One/$n$-best Decoding}

\section{Evaluation Metric}

\subsection{MRR and PMRR} % \subsection{Rank-based Metrics}
%%% Current evaluation metric for QAC: MRR, PMRR / their limitations
One of the most standard QAC evaluation metrics is the mean reciprocal rank (MRR). The MRR for the QAC system $m$ is calculated with test dataset $Q_{test}$ as follows:
\begin{equation*}
  MRR(m) = \frac{1}{|Q_{test}|} \sum_{\qv \in Q_{test}} \mathrm{RR}(\qv, m(\pv)), % \E_{p} 
\end{equation*}
where $\pv$ is a prefix of a query $\qv$ and $m(\pv)$ is the ranked list of candidate completions of $\pv$ from $m$.
$\mathrm{RR}$ is reciprocal rank of $\qv$ if $\qv$ is in $m(\pv)$, otherwise 0.

%%% Random sampling is problematic
Since it is hard to get a pair of the desired query and its prefix in a real situation, we should synthetically select a prefix by cutting off a given query for the evaluation.
Common practice is to uniformly sample from all possible prefixes within minimal length constraint in characters (or words)
% \cite{jaech2018personalized} , or include them all \cite{mitra2015query} .
% usually sampled synthetically from the given query for the evaluation based on the uniform distribution on the length.
%since we do not know the actual distribution.
%%% In each case, 
%The former 
%The latter 
However, real distribution of the prefix length may differ to the uniform distribution.
For example, users tend to engage with QAC at the position close to the boundary of words, and after typing half of query characters \cite{mitra2014user}.
Due to the stochastic characteristic of prefix sampling processes or their difference among distinct QAC systems, evaluation results are inconsistent even with the same test dataset.
To prevent this problem, a sampling function should be concretely specified.

%% PMRR
\citet{park2017neural} introduced a new metric, partial-matching MRR (PMRR):
\begin{equation*}
  PMRR(m) = \frac{1}{|Q_{test}|} \sum_{\qv \in Q_{test}} \mathrm{PRR}(\qv, m(\pv)), % \E_{p} 
\end{equation*}
where partial-matching reciprocal rank $\mathrm{PRR}$ is the reciprocal of the index of the first candidate such that the original query is the same as it or starts with it plus whitespace.
If there is no such candidate, $\mathrm{PRR}$ equals to 0. 

%Nevertheless, 
PMRR also requires sampling of the prefix length.
PMRR values are often omitted in the literature because of the similar tendency to MRR.
In other words, PMRR does not give much additional information about the quality of the generated results.

\subsection{Recoverable Length}
%%% MRL
To avoid the synthetic sampling process and length dependency, we propose a new evaluation metric for query auto-completion, namely mean recoverable length (MRL).
We define recoverable length $\mathrm{RL}$ as the number of characters right before the first position where candidates do not have the query.
When all characters of a query are known, we can readily suggest itself.
If we delete chars from right to left one-by-one, the ground truth query will disappear in the list of candidate completions.
For example, if $\qv \in m(\qv_{1:|\qv|-l})$ for $l=1, 2, 3$ but not $4$, recoverable length of this query with respect to the QAC system $m$ is 3.
MRL is mean of recoverable length:
\begin{equation*}
  MRL(m) = \frac{1}{|Q_{test}|} \sum_{\qv \in Q_{test}} \mathrm{RL}_{m}(\qv)
\end{equation*}
%MRL is postive and not upper-bounded.

%Values of MRR and PMRR do not have specific meaning so we should compare models relatively.
%We know the higher is, the better. however,...?

% Additive QAC
%PMRR: less dependency? on length but also requires sampling.
%PMRR can be useful for additive QAC, especially when it comes to mobile setting \cite{vargas2016term}. 
MRL is a useful metric for additive QAC which suggests one word at a time instead of a whole-query completion \cite{vargas2016term} in that it measures how many characters the system can predict correctly at once. % more than PMRR?
%Similiar to PMRR, MRL could be extended to partial-matching MRL (PMRL) in the future.
MRL does not care about the order of candidates and check whether they contain the target query or not.
Lastly, it eliminates the need to choose a prefix length in the test data.
%MRL does not require sampling a prefix.

%Monotonicity? (with an assumption that search is optimal)

\section{Experiments}
\begin{table*}[t!]
    \centering
    \begin{tabularx}{\textwidth}{L{0.9} *{9}{|R{0.4}} *{2}{|R{0.65}} *{1}{|R{0.4}}}
        \hline
        \multicolumn{1}{c|}{\multirow{2}{*}{\bf Model}} & \multicolumn{3}{c|}{\bf MRR}  & \multicolumn{3}{c|}{\bf PMRR} &  \multicolumn{3}{c|}{\bf MRL} & \multicolumn{2}{c|}{\bf \makecell{Execution Speed \\ \footnotesize (QPS)}} & \bf \footnotesize \multirow{2}{*}{\makecell{Decode \\ Length}} \\
        \cline{2-12}
        \multicolumn{1}{c|}{} & \footnotesize Seen & \footnotesize Unseen & \footnotesize All & \footnotesize Seen & \footnotesize Unseen & \footnotesize All & \footnotesize Seen & \footnotesize Unseen & \footnotesize All & \makecell{\footnotesize CPU} & \makecell{\footnotesize GPU} & \multicolumn{1}{c}{} \\

        \hline
        MPC                    & .570 & .000 & .290 & .616 & .095 & .360 & 8.06 & 0.00 & 4.10 &  \multicolumn{2}{c|}{$>$100} & N/A \\
        \hline
        
        \hline
        Char                    & .458 & .160 & .311 & .552 & .372 & .464 & 5.77 & 4.24 & 5.02 &  11.0 \footnotesize (1.0x) & 16.5 \footnotesize (1.0x) & 14.5 \\
        \hline
        
        \hline
        BPE                     & .242 & .085 & .164 & .305 & .232 & .269 & 0.49 & 0.54 & 0.51 & 24.2 \footnotesize (2.2x) & 37.4 \footnotesize (2.3x) & 7.1 \\
        BPE+R\tsup{1}           & .427 & .156 & .294 & .517 & .368 & .444 & {5.28} & {3.98} & {4.64} & 15.8 \footnotesize (1.4x) & 27.3 \footnotesize (1.7x) & 11.8 \\
        BPE+R\tsup{2}           & .430 & .157 & .296 & .520 & .369 & .446 & {5.44} & {4.01} & {4.74} & 15.5 \footnotesize (1.4x) & 27.2 \footnotesize (1.6x) & 12.2 \\
        BPE+R\tsup{$\infty$}    & \bf .431 & \bf .157 & \bf .296 & \bf .520 & \bf .369 & \bf .446 & \bf {5.50} & \bf {4.01} & \bf {4.76} & 15.3 \footnotesize (1.4x) & 26.9 \footnotesize (1.6x) & 12.2 \\
        \hline
        
        \hline
        SR                      & .422 & .148 & .288 & .541 & \bf \underline{.379} & .461 & 5.11 & 3.82 & 4.48 & 20.8 \footnotesize (1.9x) & 40.1 \footnotesize (2.4x) & 6.8 \\
        SR+M                    & .424 & .149 & .289 & .535 & \underline{.373} & .455 & 5.14 & 3.85 & 4.50 & 19.6 \footnotesize (1.8x) & 40.0 \footnotesize (2.4x) & 6.8 \\
        SR+R\tsup{$\infty$}     & .423 & .148 & .289 & \bf .541 & \underline{.378} & \bf .461 & 5.14 & 3.83 & 4.50 & 16.3 \footnotesize (1.5x) & 29.6 \footnotesize (1.8x) & 7.4 \\
        SR+R\tsup{$\infty$}+M   & \bf .427 & \bf .150 & \bf .291 & .538 & \underline{.375} & .458 & \bf 5.19 & \bf 3.88 & \bf 4.54 & 16.2 \footnotesize (1.5x) & 28.7 \footnotesize (1.7x) & 7.4 \\
        \hline
    \end{tabularx}
    \caption{
        \label{tab:decoding} 
        Results of completion generation. 
        We group MPC, character language model baseline, and two subword language models separately. 
        +R implies the retrace algorithm. 
        +M implies reranking with approximate marginalization.
        QPS stands for query per seconds.
        The higher the QPS, the better. The best results for each column related to accuracy are shown in \textbf{bold} for each segmentation algorithm (BPE and SR).
        SR model shows higher unseen PMRR scores (\underline{underlined}).
        Our models are faster than the character baseline.}
\end{table*}

\subsection{Data}
%%% AOL dataset and split
We use the public AOL query log dataset \cite{pass2006picture} for the experiments.
% which is the most widely used benchmark for the evaluation of QAC systems.
We split data based on time. %in own manner 
Among three months of the entire span, we use last one week as test data and one week right before the test data as validation data.
It is close to a real scenario where future queries are unseen during the training.
%Unlike random segmentation, they may come from different distributions.

%%% Preprocessing
We perform Unicode NFKC normalization and remove non-ASCII characters.
For simplicity, we change uppercase alphabets to lowercase.
After normalization and changing to lowercase, only 43 unique characters including special symbols \verb <BOS> , \verb <EOS>  and \verb <UNK>  remain.
We substituted multiple adjacent spaces to a single one and removed leading or trailing spaces.
We merged duplicates which appear adjacently by the same user and the same query.
Queries of a length shorter than three characters are filtered out. % for the test?

In total, the training, validation, test data contain 17,521,031, 1,521,971, and 1,317,632 queries, respectively.
Among the test data, 670,810 queries are seen, and 646,822 queries are unseen in the training data.
Almost half of the test data are unseen.

\subsection{Implementation Details} %\subsection{Model Architecture}
\label{implementation}
The language model used in the experiments consists of an input layer, a single  LSTM layer, a projection layer, and an output layer.
For the LSTM layer, following \citet{melis2017state} and \citet{jaech2018personalized}, we apply layer normalization \cite{ba2016layer} to each gate and couple input and forget gates.
We tie input and output embeddings for better generalization \cite{press2016using, inan2016tying}.
We set the LSTM hidden size to 600 and the input embedding size to 100.

We train three individual language models: namely Char, BPE, SR.
The only difference among models is how to segment a given sentence into tokens.
% tested combinations of input embedding size (25, 100) and hidden size (300: small, 600: large). 
%An LSTM layer takes a majority of overall model parameters.
%% As the vocabulary size increases the required dimension size to represent each token increases.
We believe that increasing model size (number of LSTM layers, input size, and hidden size) would improve the performance. 
Also, the best set of a combination may differ depending on models.
However, we use the same model size for the character baseline and our variants for the fairness since our goal is proposing a new method and comparing between baseline and ours rather than achieving the best performance with the restriction of having a similar number of parameters.
%%%%%%%%%%
% We use various vocabulary sizes of \{256, 512, 1024\} for the subword segmentation.
% Overall, we trained seven language models: namely Char, BPE\tsub{S}, BPE\tsub{M}, BPE\tsub{L}, SR\tsub{S}, SR\tsub{M}, SR\tsub{L}.
% S, M, L corresponds to a vocabulary size of 256, 512, 1024, respectively.

We use the off-the-shelf SentencePiece \cite{kudo2018sentencepiece} library for vocabulary learning and segmentation of the vocabulary size 256 using BPE, SR algorithms. % unigram language model.
For the subword regularization, we use sampling parameters $l=\infty$, $\alpha=0.2$ for the training.
% dependence on SR parameters: we fix to l:inf, alpha:0.2
We choose this value by the generation accuracy on the validation data.
Increase of model size and computation due to larger vocabulary size are not substantial.
By setting a manageable amount of vocabulary size, we can balance performance and computational cost.

For the computational efficiency, we truncated queries in the training data to a length of 40. 
% characters.
Only less than 3\% of queries in the training data are longer than 40 characters.
We train models for thirty epochs by the Adam \cite{kingma2014adam} optimizer with a learning rate 5e-3 and batch size 1024.
Following \citet{smith2018disciplined}, we use a large learning rate and batch size.
%It allows faster training until the final convergence.
We use recurrent dropout \cite{semeniuta2016recurrent} with probability of 0.25 for regularization.
The best model is chosen using validation data.

Using QAC models, we generate $N = 10$ completion candidates using beam search decoding of a beam width $B = 30$.
%which is chosen based on the validation data.
%For the beam search decoding to generate completions, 100 and a branching factor of 4.

For the SR models, the segmentation of $\pv$ (or retraced $\pv_{1:|\pv|-r}$) is not deterministic and generated completions may differ depending on its segmented token sequences with their different encoded representation.
By following \cite{kudo2018subword}, we can find the most likely segmentation sequence $\tv$ starting from all of the $n$-best segmentations $\tilde{\tv}_1, \cdots, \tilde{\tv}_n$ of $S(\pv)$ rather than from only $\tilde{\tv}_1$.
However, we observe that this $n$-best decoding performs worse than one-best decoding.
One possible reason is that segmentations which are not the best have a smaller probability as itself and so less likely to appear in training and less competitive in the process of beam search.
For this reason, we set $n$ to 1.
% For this reason, we did not report those values. % in Table ~\ref{tab:decode}.

We used a trie \cite{fredkin1960trie} data structure to implement most popular completion baseline. % frequency based 

All experiments were performed on NAVER Smart Machine Learning (NSML) platform \cite{sung2017nsml, kim2018nsml}.

\subsection{Decoding Results}
%\input{tab_decode}

%%% Explanation of variants
We performed comprehensive experiments to analyze the performance of query auto-completion.
Table~\ref{tab:decoding} shows the generation result of MPC, the character baseline, and our model variants.
% We use the character baseline and variants of subword language models mentioned in Section \ref{implementation} for decoding.
For BPE models, we varied the maximum retrace step to 0 (without retrace algorithm), 1, 2, and $\infty$ (no limitation on retracing step size).
For SR models, we compare decoding results without any techniques, with marginalization only, with retrace algorithm only, and with both. %
%We will add a table containing all of these results to the appendix. For the visibility, the representatives are chosen and put to Figure \ref{fig:decoding}.

MPC is a very fast and remarkably strong baseline.
It is worse than language models in the overall score (MRR, PMRR, and MRL), but better for previously seen queries.
However, it is unable to predict unseen queries.
Even with efficient data structures, MPC requires huge memory to keep statistics of all previous queries.
As a practical view, combining frequency-based traditional method and neural language model approach can boost the accuracy and meet trade-off between the performance and computational costs.

%Trade off between MRR and PMRR?
%%% MRR / PMRR  
% Models of small vocabulary size with retrace algorithms and marginalization performs best for both BPE and SR.
MRRs and PMRRs of our best methods are close to that of the character model with less than 0.02 point drop.
%%% Unseen PMRR -> better generalization
%Subword models achieve much higher PMRR than the character baseline. -> No
Notably, the SR model has better generalization ability in that their PMRR for unseen queries is higher than that of the character model.
In a real scenario, it is more critical because unseen queries come in increasingly as time goes by.
%%% $n$-best decoding

%%% Time: Execution speed, Decoding length
% Execution speed is the number of queries processed every second (QPS).
We measure execution time with Tesla P40 GPU and Xeon CPU.
Subword-level models are up to 2.5 times faster than the character baseline with minimal loss in performance both in CPU and GPU.
Decoding length which is maximum suffix length until beam search ends correlates with the number of floating-point operations.
Subword models significantly reduce the decoding length from the character baseline more than two times shorter by generating multiple characters at once.
%Decoding length of BPE is smaller than SR due to the deterministic nature of segmentation, resulting in much faster execution speed. -> No

%%% Retrace algorithm
Models with additional techniques perform better than without them.
Especially, retrace algorithm gives huge improvement for BPE case.
Without retrace algorithm, BPE models do not work well.
On the other hand, SR models only obtain small improvement.
%BPE models with the retrace algorithm and all SR models achieve much higher PMRR than the character baseline.
Because retrace algorithm goes back, it increases the decoding length and slows down the speed. % about 1-2  %  a little
Although current retrace algorithm is implemented straightforwardly, it can be improved by merging beams efficiently.
Most of the subword lengths are equal or shorter than 3, so retrace of step 2 is quite enough, and R\tsup{2} get a close result with R\tsup{$\infty$}.
%By changing this limit, we observe -.

%%% Reranking by approximate marginalization
The reranking method by approximate marginalization gives a small amount of improvement and is orthogonal to retrace algorithm.
Marginalization method increases MRR but decreases PMRR.
It is plausible in that it changes the order of candidates by reranking.
The effect of marginalization would be better if we use a bigger beam size.
Because the reranking process is done after beam search decoding which takes most of the decoding time and only consists of summation and sorting the final beam outputs, it does not take a long time.

%%% vocab size
We also had experimented by increasing the vocabulary size.
The accuracy of BPE models degrades fast as the vocabulary size increases.
On the other hand, the performance of SR models is quite stable due to the regularization effect during training.
As desired, the larger the dictionary size, the shorter the decoding length.
%We can use a larger vocabulary size if faster decoding is needed. 
Whereas computations run in parallel in GPU, the number of operations for the output layer in the language model is proportional to the vocabulary size in CPU.
Therefore, a larger vocabulary size does not always guarantee speedup for execution in the CPU.
More thorough investigation about the correlation between QAC performance and the vocabulary size of subword language models remains for future work.

%\subsection{Qualitative Analysis}
\begin{table}[t!]
    \small
    \centering
    \begin{tabular}{ll|ll}
      %\hline
      %\multicolumn{2}{c}{re}\\ %  & \multicolumn{2}{c}{nat} 
      \hline
      \multicolumn{2}{c|}{\bf re} & \multicolumn{2}{c}{\bf nat}  \\
      \hline
      \multicolumn{1}{c}{\small Char} & \multicolumn{1}{c|}{\small SR} & 
      \multicolumn{1}{c}{\small Char} & \multicolumn{1}{c}{\small SR} \\
      %\tsub{S}+R\tsup{$\infty$}+M
      \hline
      \scriptsize realtor.com   & \scriptsize recipes       & \scriptsize national city bank    & \scriptsize national bank\\
      \scriptsize recipes       & \scriptsize rentals       & \scriptsize nationalcity.com      & \scriptsize national city\\
      \scriptsize real estate   & \scriptsize real estate   & \scriptsize national city         & \scriptsize nationwide\\
      \scriptsize remax         & \scriptsize restaurants   & \scriptsize national geographic   & \scriptsize national parks \\
      \scriptsize realtor       & \scriptsize resources     & \scriptsize national car rental   & \scriptsize national park \\
    
      \hline
    \end{tabular}
    \caption{\label{tab:example} Examples of top 5 candidates of completions given "re" and "nat" as prefixes generaed by the character baseline and SR model.}
\end{table}

\begin{figure*}[t!]
  \centering
  \includegraphics[width=\textwidth]{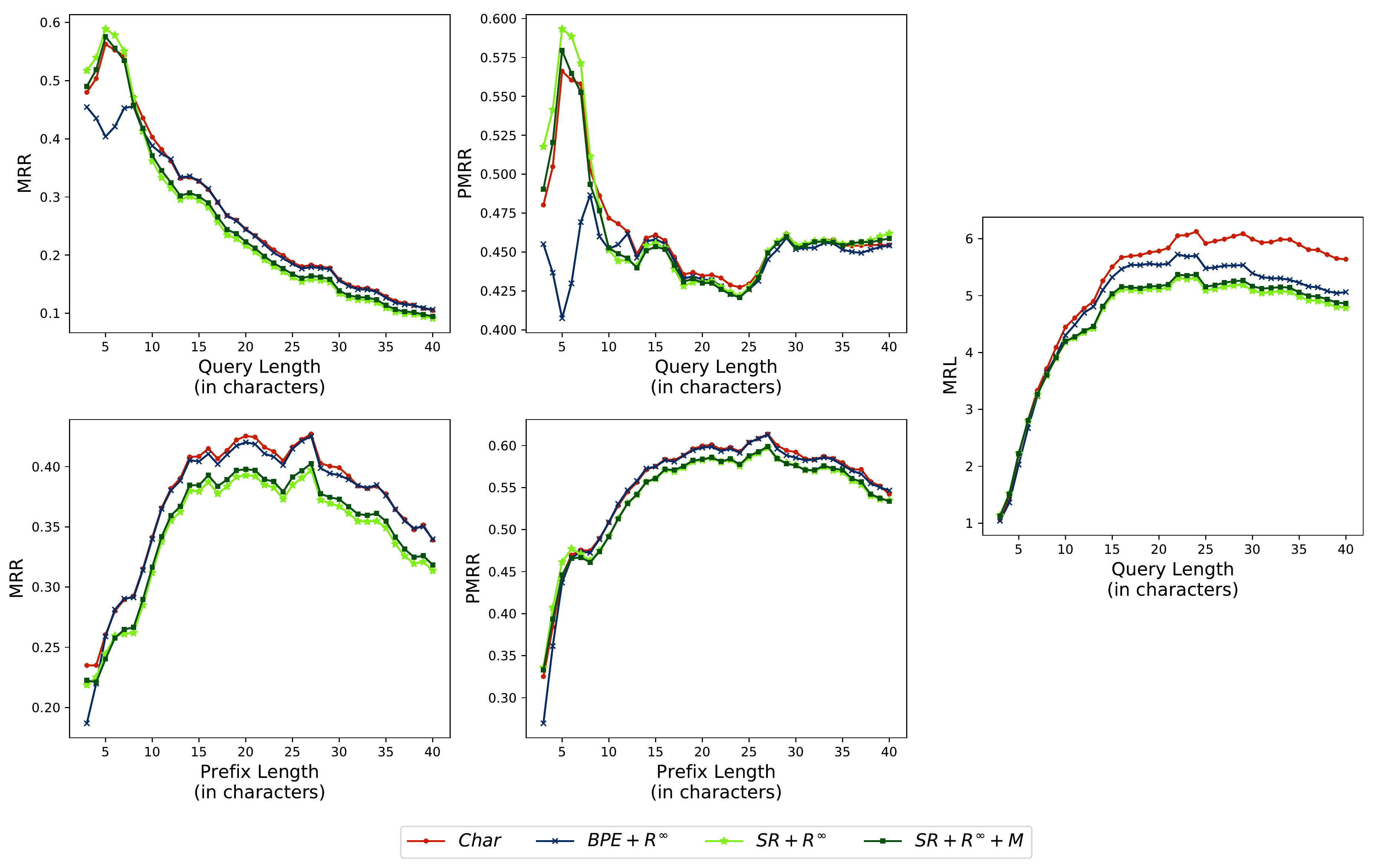}
  \caption{Comparison of the character-level baseline model and our best models by changing query length and prefix length in terms of three evaluation metrics: MRR, PMRR, and MRL. MRL is only varied by query length because it does not require prefix length sampling.}
  \label{fig:metric}
\end{figure*}

Table \ref{tab:example} shows examples of decoding results.
Our model generates more concise and realistic examples.

\subsection{Analysis on Evaluation Metrics}

As shown in Figure~\ref{fig:metric}, we compared our best models on three evaluation metrics (MRR, PMRR, and MRL) by changing the query length and prefix length.
MRR and PMRR are more favorable to shorter queries.
They drop very fast as the query becomes longer. % as seen in Figure 
%Same with MRR, PMRR tends to have a higher value for short queries.
% of auto-regressive models usually tend to generate short completions.
%even with length normalization \cite{wu2016google},
For a longer query, the suffix length after sampling prefix has more chance to be longer.
The search space increases exponentially with its suffix length.
Even though QAC systems could generate realistic candidates, it is quite hard to match a long sequence with the ground truth.
% Two methods of having similar MRR values might perform differently in reality.
% In this sense, MRR is not good enough to evaluate the results of query auto-completion.
As the prefix length becomes longer which means that much information for determining the query has been given, the completion performance improves.

Interestingly, MRR and MRL of BPE are higher than those of SR, although BPE is worse in terms of PMRR than SR.
For short queries, SR outperforms the character baseline.
On the other hand, BPE is poor when the query length (or prefix length) is short.
However, for a longer case, its MRR is almost close to that of the character baseline.

%%% MRL
%Our best methods also achieve similar MRL with the character baseline (5.02): 4.76 and 4.54 for BPE and SR, respectively, with the retrace algorithm and reranking method.
MRR and PMRR are highly dependent on the length distribution of test data.
% Moreover, as shown in Table \ref{tab:decode}, MRL is more fine-grained.
In contrast, MRL keeps the order between different methods as the query length changes. %, but other metrics do not.
MRL is more reliable in the respect that it could provide consistent order between methods regardless of query length distribution.
% we have observed that MRL is more fine-grained than MRR and PMRR.
% Length dependency
%On the other hand, PMRR and MRL are less dependent on the query's length.
For long queries lengths, MRL stays in the flat area.
Normalizing recoverable length based on the query length might be necessary. %and is an open problem.

\section{Future Work}
%%% Future work
Approximation in training (Section \ref{sec:training}) and decoding (Section \ref{sec:decoding}) deteriorate the accuracy of subword language modeling.
One possible solution to reduce the accuracy gap between the character language model baseline and the subword language model is knowledge distillation \cite{hinton2015distilling, kim2016sequence, liu2018distilling} from character-level language models.
A student model can learn to match an estimation of query probability with that of a teacher model.

%%% marginalization
Another interesting research direction is learning segmentation jointly with language model \cite{kawakami2019learning, grave2019training} rather than using fixed pretrained segmentation algorithms.
%(2) constructing a subword vocabulary in other ways including learning optimal vocabulary by training jointly with a language model,
A conditional semi-Markov assumption allows exact marginalization using dynamic programming \cite{ling2016latent, wang2017sequence}.
Nevertheless, beam search decoding on those language models, especially faster decoding, is non-trivial.

%%% Beam search decoding
%%%%%%%%%For better decoding algorithms considering such as length normalization \cite{wu2016google} and diversity \cite{cai2016diversifying, li2016simple, vijayakumar2016diverse}.
% noisy model \cite{xie2017data}, 
%optimizing bsd
% SR
% decoding together with retraced one

Proposed method can be extended to wide range of tasks.
Query suggestion \cite{sordoni2015hierarchical, dehghani2017learning} and query reformulation \cite{nogueira2017task} are related to QAC and well-established problems.
They both are also possible applications of the subword-level modeling.
%\citet{sordoni2015hierarchical} and \citet{dehghani2017learning} leverage a word-level hierarchical recurrent encoder-decoder and a copy mechanism \cite{vinyals2015pointer, see2017get} for query suggestion.
%\citet{wang2018realtime} incorporates a noisy model for error correction of QAC.
\cite{drexler2019subword} used subword regularization and beam search decoding for end-to-end automatic speech recognition.

% Hybrid with MPC and lambdaMart
Lastly, implementation with more advanced data structure \cite{hsu2013space} and parallel algorithms to speed up and meet memory limitation are necessary for the real deployment \cite{wang2018realtime}. It would be helpful if the computation is adaptively controllable on-the-fly \cite{graves2016adaptive} at the runtime depending on the situation.

\section{Conclusion}
%%% Contribution
In this paper, we propose subword language models for query auto-completion with additional techniques, retrace algorithm and reranking with approximate marginalization.
%for fast and accurate candidate generation.
We observed subword language models significant speedup compared to the character-level baseline while maintaining the generation quality.
Our best models achieve up to 2.5 times faster decoding speed with less than 0.02 point drop of MRR and PMRR. 

Using a subword language model, we build an accurate and much faster QAC system compared to the character-level language model baseline.
Although there is still much room for improvement on hyperparameter optimization, decoding search, and neural architectures like Transformer \cite{vaswani2017attention, dai2019transformer}, the goal of this work is to prove that the subword language model is an attractive choice for QAC as an alternative to the character-level language model, especially if latency is considered.

We believe that our newly proposed metric, mean recoverable length (MRL), provides fruitful information for the QAC research in addition to conventional evaluation metric based on ranks.

\section*{Acknowledgments}
The author would like to thank Clova AI members and the anonymous reviewers for their helpful comments.
% for proofreading this manuscript.

\bibliographystyle{acl_natbib}
\bibliography{ms}

%\appendix

\end{document}